\documentclass[10pt,twocolumn,letterpaper]{article}

\usepackage{wacv}
\usepackage{times}
\usepackage{epsfig}
\usepackage{graphicx}
\usepackage{amsmath}
\usepackage{amssymb}

\def\wacvPaperID{491} 

\wacvfinalcopy 

\ifwacvfinal
\def\assignedStartPage{1} 
\fi

\ifwacvfinal
\usepackage[breaklinks=true,bookmarks=false]{hyperref}
\else
\usepackage[pagebackref=true,breaklinks=true,colorlinks,bookmarks=false]{hyperref}
\fi

\ifwacvfinal
\else
\fi

\newcommand{\argmax}{\operatornamewithlimits{argmax}}
\begin{document}

\title{Separable Four Points Fundamental Matrix}

\author{Gil Ben-Artzi\\
Computer Science Department\\
Ariel University\\
{\tt\small gilba@g.ariel.ac.il}
}

\maketitle

\begin{abstract}
We present a novel approach for RANSAC-based computation of the fundamental matrix based on epipolar homography decomposition. We analyze the geometrical meaning of the decomposition-based representation and show that it directly induces a consecutive sampling strategy of two independent sets of correspondences. We show that our method guarantees a minimal number of evaluated hypotheses with respect to current minimal approaches, on the condition that there are four correspondences on an image line. We validate our approach on real-world image pairs, providing fast and accurate results.
\end{abstract}

\section{Introduction}
\label{Sec:intro}
One of the basic building blocks in computer vision is the computation of epipolar geometry given a set of putative image point correspondences. Often, such correspondences include mismatches, therefore a robust estimation method needs to be carried out. The most common method used is RANSAC~\cite{Fischler}. RANSAC-based computation of the fundamental matrix iteratively samples a minimal set of putative correspondences and hypothesizes the fundamental matrix parameters. The goodness-of-fit of each hypothesis is evaluated with respect to the whole set of the putative correspondences. The number of evaluations at each iteration is between one to three, depending on the number of possible solutions. This process is repeated until a predefined number of iterations is exceeded.  One of its key limitations is that in the presence of a considerable amount of mismatched points' correspondences, a large number of hypotheses' evaluations is needed in order to obtain a reliable model. The number of the required hypotheses is directly related to the number of points needed to be sampled in order to hypothesize the fundamental matrix parameters. From a geometrical point of view, the minimal sample size is at least seven \cite{hartley2003multiple}. Minimal sample approaches have been proposed requiring only six \cite{wang1997six} or five points' correspondences \cite{stewenius2005minimal,barath2018five}, with additional knowledge regarding the parameters or the scene.

\begin{figure}[t]
\begin{center}
\begin{tabular}{cc}
 \includegraphics[height=0.47\linewidth]{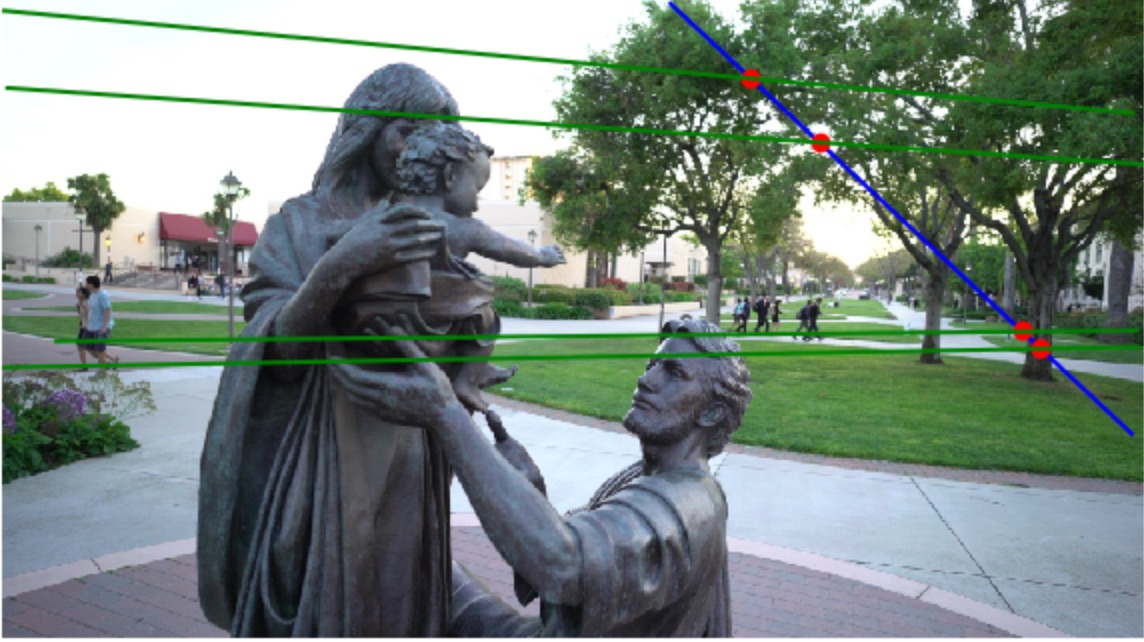}\\
 \includegraphics[height=0.47\linewidth]{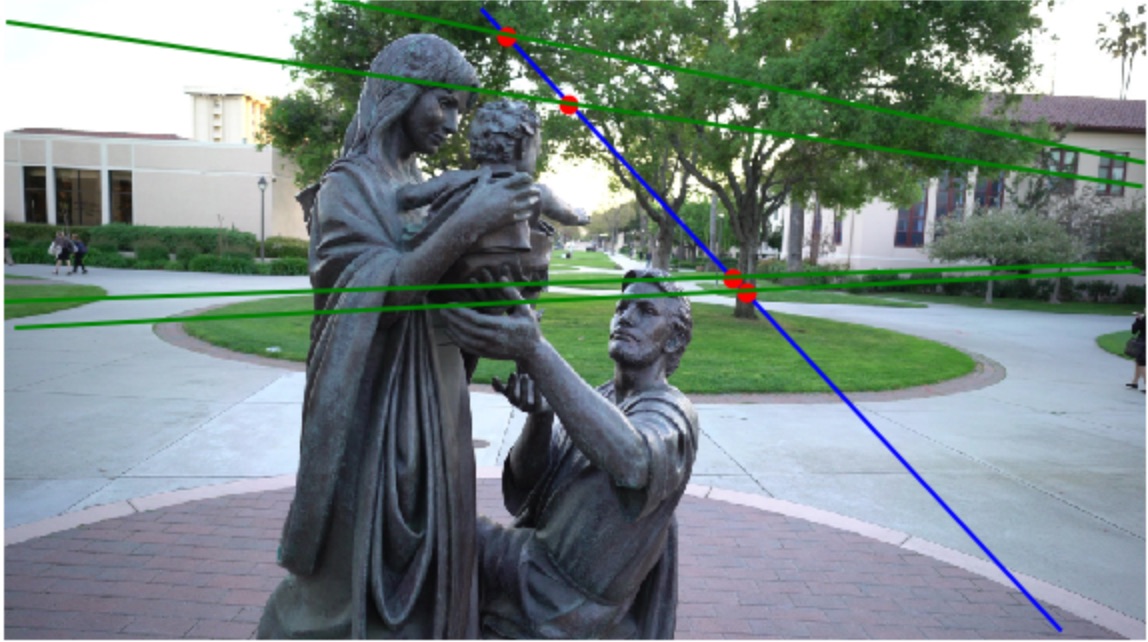}
\end{tabular}
\end{center}
  \caption{Our approach matches points on line segments in the images without the need for the full computation of the fundamental matrix. Based on the matched points and line segments, we use our reformulation to compute the epipolar homography and reduce the required number of RANSAC iterations for the full computation. The red points are the matched putative point correspondences which are aligned on line segments (blue lines) in the image planes. The green lines are epipolar lines which are not known a-priori and do not take part in the matching process. The matching process is based on the computation of the epipolar homography which, unlike commonly used methods, does not require the knowledge of epipolar line correspondences.}
\label{fig:open}
\end{figure}

We present a method that can reduce the minimal number of RANSAC hypotheses' evaluations beyond the current known minimum. Our approach, similar to approaches that require less than seven correspondences, assumes additional knowledge. We require that a single line segment exists in the image with four correspondences and show experimentally that such a configuration is common. Given such a configuration, we show that our approach guarantees a minimal number of hypotheses' evaluations. In cases where such a configuration does not exist, standard RANSAC can be applied with overall small overhead.  Hereafter, when referring to RANSAC iterations, we will refer to the actual number of hypotheses' evaluations which depends upon the number of solutions per hypothesis.

We compute the fundamental matrix in two steps. In the first step, we sample three points on a line segment. We compute the epipolar homography and validate its goodness of fit using at least one additional point, without the full computation of the fundamental matrix. In the second step, the recovered matches are fixed and the remaining points' matches are recovered by sampling, without any restriction.  Our approach is based on the computation of the epipolar homography. Existing methods for the computation of the epipolar homography~\cite{kasten2016fundamental,sinha2010camera,ben2016camera,halperin2018epipolar,hartley2003multiple} require the knowledge of (at least) three corresponding epipolar lines and, as a consequence, are rarely used in practice. We show how to compute the epipolar homography without a-priori knowledge of epipolar lines.  Figure~\ref{fig:open} shows the recovered four matches in the first step of our approach and the extracted epipolar lines based on our epipolar homography computation.

This paper therefore contributes by presenting: (a) a reformulation of the fundamental matrix based on epipolar homography decomposition and an analysis of its geometrical meaning, (b) a novel two steps RANSAC-based approach for the computation of the fundamental matrix that can markedly reduce the required number of hypotheses' evaluations and (c) an efficient implementation and a validation of our approach on real-world image pairs showing that it is accurate and applicable.

\section{Related Work}

The most common approaches for the computation of the fundamental matrix are the seven or eight point algorithms~\cite{hartley2003multiple}. The eight-point algorithm~\cite{longuet1987computer} which was adapted for the fundamental matrix, was made practical by \cite{hartley1997defense}. It is based on the normalization of the points' correspondences, and computes the parameters based on the Direct Linear Transform (DLT) by enforcing the rank 2 constraint \cite{luong1996fundamental}. The seven points algorithm relaxes the eight points requirement by an additional zero determinant constraint of the matrix, resulting in a cubic equation with one or three real solutions.

Various minimal methods assume additional knowledge in order to reduce the required number of correspondences. If the camera parameters are known, the five-points algorithm \cite{nister2004efficient} can be used. Recently, Barath \cite{barath2018five} proposed an approach for the estimation of the fundamental matrix based on five correspondences, in cases where three co-planar point correspondences and the rotation of features are known. Ben-Artzi et al. \cite{ben2016epipolar} showed that in case the intensity of the epipolar lines are similar across views, two corresponding points are sufficient for the computation of matching epipolar lines, and can be used for recovery of the fundamental matrix. 

Methods attempting to empirically reduce the number of RANSAC iterations have also been introduced \cite{barath2019magsac,chum2005matching,nister2005preemptive}. Unlike such methods, our approach, as well as other minimal methods, guarantees to reduce the number of required samples and this number can be calculated in advance in the same way as in standard RANSAC. In addition, our approach has a true geometrical meaning (See Sec.~\ref{Sec:Theo}), whereas no geometrical analysis is possible in such methods.

The fundamental matrix can be computed based on the epipolar homography ~\cite{hartley2003multiple,faugeras2001geometry}. These methods require the knowledge of epipolar line correspondences. Sinha~\cite{sinha2010camera} introduced an approach for camera calibration based on the computation of the epipolar homography and epipoles hypothesizing. His approach was later improved by \cite{ben2016camera}, who directly recovered the epipolar lines required for the computation of the epipolar homography by using the motion-barcode descriptor. Similar approaches were introduced by ~\cite{kasten2016fundamental,halperin2018epipolar} for the computation of the epipolar homography by directly estimating the epipolar lines. However, all of the above methods are only applicable to videos of dynamic scenes. Wurfl et. al. \cite{wurfl2019estimating} presented an approach for estimating the fundamental matrix transmission imaging, based on the epipolar homography formulation. They do not require explicit correspondences but rather sampling of all edge pixels of the image. The computation of the relative pose by general line homography based on line segments was demonstrated by \cite{reisner20103d} but they required the matching points to be the projection of existing 3D lines in the scene and required at least two such lines.

\section{Theoretical Background}
\label{Sec:Theo}

{\bf Epipolar-Lines Based Parametrization.} Let $l,l'$ be corresponding epipolar lines, in the first and second image, respectively.  Let $l_p$ be a line that intersects $l$ in the point $p$ and does not include the epipole. It follows that $p$ = $l_p \times  l$, where $\times $ is the cross product. Since $p$ is on $l$, it follows that $l^{'} = F p$.  Denote $[\cdot]_{\text{x}}$ as the skew symmetric matrix associated with the cross product, we have the following mapping:

$$
l^{'} = F [l_p]_{\text{x}} l
$$

where $H_l=F [l_p]_{\text{x}}$ is the epipolar homography. The seven degrees of freedom of the fundamental matrix is constructed by the four degrees of freedom of the two epipoles and the three degrees of freedom of the epipolar homography \cite{hartley2003multiple}. The epipolar homography is obtained by three corresponding epipolar lines.\\

\begin{figure}[tb]
\begin{center}
\includegraphics[width=0.67\linewidth]{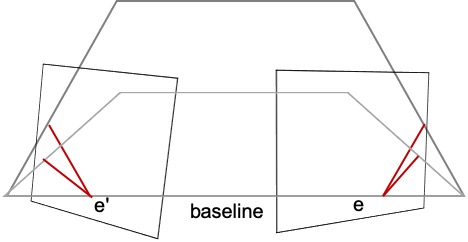}
\end{center}
\caption{The pencil of epipolar planes. As the epipolar lines (red lines) rotate around the epipoles (e,e'), their associated planes rotate around the baseline.}
\label{fig:plane_penci}
\end{figure}

{\bf Image-Points' Based Parametrization.} Consider the set of epipolar lines passing through the epipole which is denoted as the \emph{pencil of epipolar lines}. As shown in Fig.~\ref{fig:plane_penci}, the corresponding epipolar lines across images define a plane which intersects the two retinal planes with its axis as the baseline. When the corresponding epipolar lines rotate around the epipoles, their corresponding planes rotate around the baseline, defining the \emph{pencil of epipolar planes}. The two corresponding pencils of epipolar lines are related by the epipolar homography based on the pencil of epipolar planes \cite{faugeras2001geometry}. Let  $(l_i,l^{'}_i)$ be corresponding epipolar lines in the first and second image respectively. Let $\bar{l},\bar{l}'$ be arbitrary lines in the first and second image, such that the epipoles $e,e'$ are not on the lines. The $2 \times 2$ epipolar homography $H_e$ maps between $(l_i,l^{'}_i)$ based on their intersection \emph{points} with  $\bar{l},\bar{l}'$ as follows.  Let $x_1,x_2 \in \bar{l}$ and $x_1',x_2' \in \bar{l}'$, denoted as \emph{control points}. The intersection points of $(l_i,l^{'}_i)$ with $\bar{l},\bar{l}'$ are $x_{p}=(x_1 \times x_2) \times l_i$ and  $x_{p'}=(x_1' \times x_2') \times l_i'$, where $\times$ is the cross product and using  homogeneous coordinates of the control points from now on. Using the following identity
$$
v ~~\times ~~ (u ~~\times~~ w) = (v^T w) u + (-v^T u) w 
$$

for $u,v,w \in \mathbb{R}^{3\times1}$, the intersection points can be written as a linear combination of the control points as $x_{p}=\alpha x_1 + \beta x_2$ and $x_{p'}=\alpha' x_1'+\beta' x_2'$, where 
$$
\alpha=l_i^T x_2,~~\beta=-l_i^T x_1,~~\alpha'=l_i'^T x_2',~~\beta'=-l_i'^T x_1',
$$

${\alpha,\beta,\alpha',\beta'} \in R$.  The $2 \times 2$ epipolar homography is given by:\\

\begin{equation}
\begin{bmatrix} 
\alpha' \\
\beta'
\end{bmatrix}
=
\underbrace{
\begin{bmatrix} 
a & b \\
c & d 
\end{bmatrix}
}_{H_e}
\begin{bmatrix} 
\alpha \\
\beta
\end{bmatrix}
\end{equation}

\begin{figure}[tb]
\begin{center}
\includegraphics[width=0.98\linewidth]{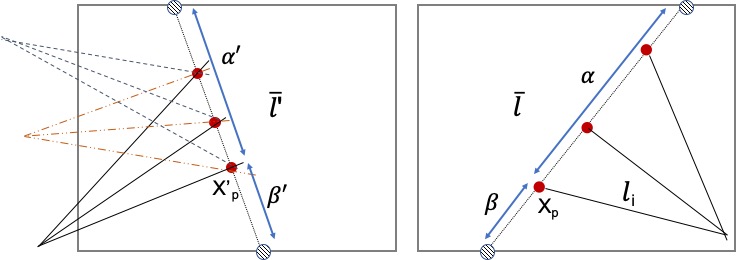}
\end{center}
\caption{ The epipolar homography has three degrees of freedoms.  Existing methods compute the epipolar homography based on epipolar lines. The points-based parametrization enables the epipolar homography's computation regardless of the epipole positions. On the left image, various epipole positions are valid for the same epipolar homography. The parametrization of the intersection point $x_p$ of the epipolar line $l_i$ is based on a linear combination $\alpha,\beta$ of selected fixed points on the lines, marked by the striped points, intersecting the boundaries of the image. $x_p'$ is the corresponding intersection point of $x_p$ and is parametrized by $\bar{l}', \alpha',\beta'$. See text for details.}
\label{fig:epipolar_hom}
\end{figure}

~~~~~~~~\\

Rewriting the corresponding epipolar lines as the cross product of the epipoles and points on the epipolar lines, and expanding the intersection points' coefficients, we have the following parametrization of the fundamental matrix:

\begin{equation}
F \approx
\begin{bmatrix} 
e_2
\end{bmatrix}_{\text{x}}
\begin{bmatrix} 
x_1',x_2'
\end{bmatrix}
\begin{bmatrix} 
a & b \\
c & d 
\end{bmatrix}
\begin{bmatrix} 
x_2^T\\
-x_1^T
\end{bmatrix}
\begin{bmatrix} 
e_1
\end{bmatrix}_{\text{x}}
\label{eq:H}
\end{equation}

~~~~~~~~\\

{\bf Observation 1.} The epipolar homography $H_e$ maps between the corresponding points $x_p,x_p'$ based on their representation as a linear combination of their respective control points. It can be computed without knowledge of the corresponding epipolar lines. Given the line segments of the points, we select (fixed) control points on the lines and represent the points as their linear combination. The points-based parametrization of the pencil of epipolar lines by their intersection with the lines $\bar{l},\bar{l}'$ is shown in Fig.~\ref{fig:epipolar_hom}.\\

\begin{figure}[t]
\begin{center}
\includegraphics[height=0.7\linewidth]{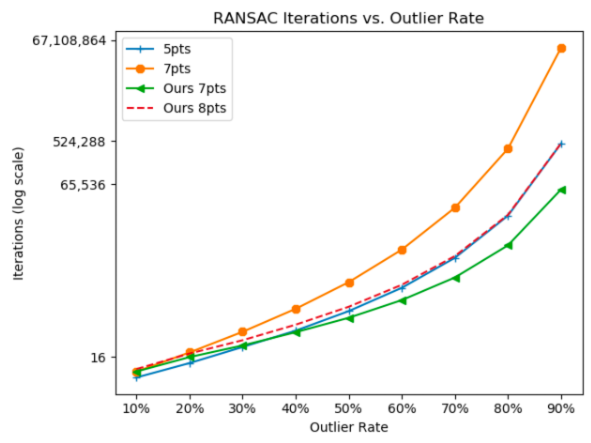} 
\end{center}
   \caption{ The theoretical number of RANSAC iterations for the computation of the fundamental matrix by minimal methods. 7pts represents the seven-point algorithm and 5pts represents a RANSAC-based approach requiring a sample size of five (e.g. \cite{barath2018five}). Ours-7pts represents our approach based on the seven-point algorithm and Ours-8pts represents our approach based on the eight-point algorithm.}
\label{fig:ransac_iterations_theoretical}
\end{figure}

{\bf Observation 2.} The required number of RANSAC iterations for the full computation of the fundamental matrix can be greatly reduced. First, we recover $H_e$ which is a three parameters model without the need for the full computation of the fundamental matrix. We can then recover four additional arbitrary correspondences by relying on the recovered $H_e$. Let $F(k,r)$ be the required number of RANSAC iterations for a model with $k$ parameters, outlier rate $r$ and success probability of $0.99$. The required number of RANSAC iterations is now $F(3,r)+F(4,r)$ instead of $F(7,r)$. Fig.~\ref{fig:ransac_iterations_theoretical} shows the required number of RANSAC iterations for the computation of the fundamental matrix by our approach and existing minimal methods, as a function of the outlier rate. Our approach can be implemented based on the seven-point or eight-point algorithm; both are presented in the figure. The number of RANSAC iterations is computed assuming one solution per sample and no additional overhead for any of the methods. In Sec.~\ref{Sec:exp}, we present the expected number of RANSAC iterations based on the actual number of solutions per sample in the tested datasets. \\


\begin{figure}[tb]
\begin{center}
 \includegraphics[width=0.95\linewidth]{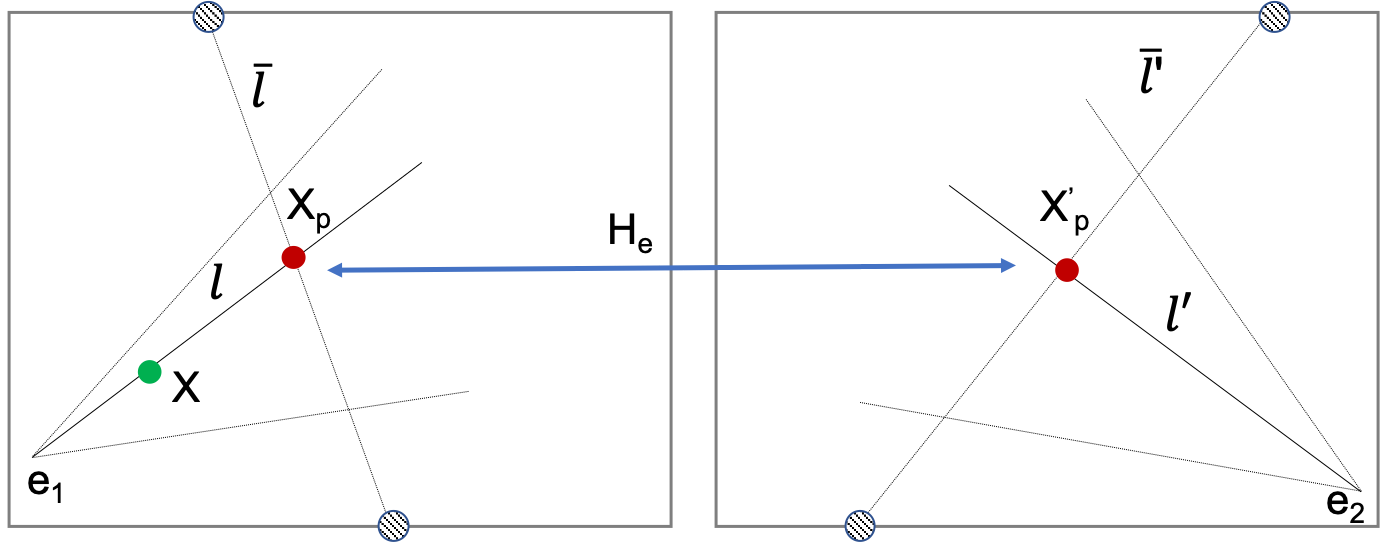}
 \end{center}
   \caption{The mapping of a point on the left image to the corresponding epipolar line on the right image by the fundamental matrix, parametrized by points-based epipolar homography $H_e$: (a) from a point $x$ to epipolar line $l$ based on the known epipole $e_1$, (b) from the epipolar line $l$ to the intersection point $x_p$ with a fixed arbitrary image line $\bar{l}$, (c) from the intersection point $x_p$ to the corresponding intersection point $x_p'$ with a fixed arbitrary line $\bar{l}'$ and (d) from the point of intersection $x_p'$  to the epipolar line $l^{'}$ based on the known epipole $e_2$.}
\label{fig:hom_mapping}
\end{figure}

{\bf Observation 3.} The lines $\bar{l},\bar{l}'$ with the corresponding points are arbitrary line segments in the images.  Figure~\ref{fig:hom_mapping} illustrates the mapping of points in the first image to lines in the second image, based on the parametrization of the fundamental matrix using the points-based epipolar homography (Eq.~\ref{eq:H}), given $\bar{l},\bar{l}'$.

\section{Line Segments with Corresponding Points}
\label{Sec:Lines}

\begin{figure}[tb]
\begin{center}
   \includegraphics[width=1\linewidth]{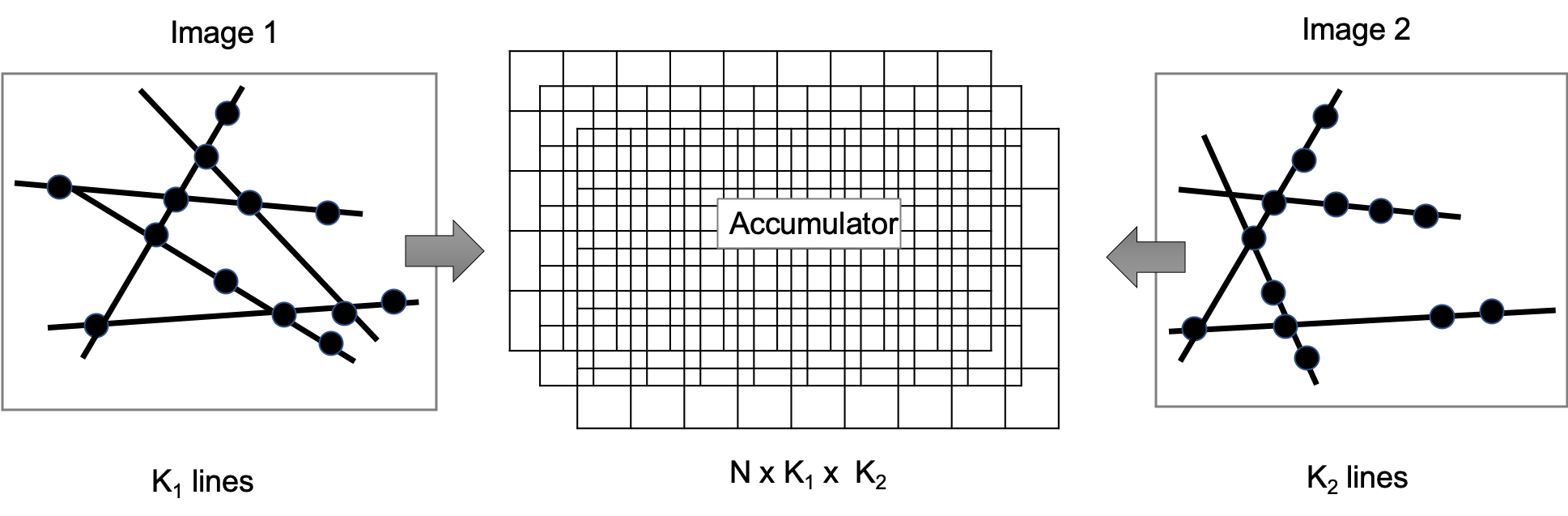}
\end{center}
   \caption{We match points on line segments without the need to explicitly compare between lines across images. In each image, Hough transform is carried out and a shared accumulator of size  $N \times K_1 \times K_2$ is updated independently, where $N$ is the number of correspondences, $K_1$ is the number of lines in the first image and $K_2$ is the number of lines in the second image.   }
\label{fig:match_lines}
\end{figure}

In this section we describe how to find the lines $\bar{l},\bar{l}'$ across images with the maximum number of putative correspondences. For each image, our approach does not require knowledge of the points and lines in the other image, which allows parallel implementation. The key idea is to use Hough transform \cite{duda1971use} to detect the lines independently in each image and update a joint accumulator which can be queried only after the process is completed. Fig.~\ref{fig:match_lines} illustrates the matching process of corresponding lines across images.  

Let $I_1,I_2$ denote a pair of images. Let  $X=\{(x_i,x_i')\}_{i=1..N}$ denote the set of ordered pairs of unique \emph{putative corresponding image points} where $x_i,x_i' \in \mathbb{R}^2$ are image points in $I_1,I_2$, respectively. Let $\bar{X}=\{x_i\}_{i=1}^N$, $\bar{X'}=\{x_i'\}_{i=1}^N$. We create binary images $B_1,B_2$ such that for each pixel $y_i$ in $I_1$, $B_1(y_i)=1$ if $y_i \in \bar{X}$ otherwise $B_1(y_i)=0$, and similarly for $B^{'}_2$. We use Hough transform \cite{duda1971use} to extract the ordered set of lines $L_1=\{l_i\}_{i=1}^{K_1},L_2=\{l_i'\}_{i=1}^{K_2}$ in $B_1,B_2$, respectively. We consider only lines with at least four correspondences. This requirement is due to the robust estimation process of the homography (See Sec.~\ref{Sec:Sep}), where we sample three points and need at least one additional point to estimate the quality of the hypothesized homography.

We define a multidimensional array, an accumulator $A$ of  size $N \times K_1 \times K_2$, where  $K_1$ and $K_2$  are the number of lines in $L_1$ and $L_2$, respectively, and $N$ is the number of putative corresponding points. The accumulator is initialized to zero. Let $D_1(j) \subset \{1...K_1 \}$ be the indices of nearby lines in $L_1$ for a point $x_j$ in image $I_1$,

\begin{equation}
D_1(j)=\big\{i| l_i \in L_1,  x_j \in \bar{X}, d(x_j,l_i) < C \big\},
\end{equation}
where $d$ is the point to line distance and $C$ is the constant representing the required distance between the lines and points. For each point $x_j$ in the first image, we increment the accumulator according to its nearby lines in the first image and all existing lines in the second image, $A(j,m,n)+=1$ s.t. 

\begin{equation}
\big\{(j,m,n)| m \in D_1(j), n \in \{1...K_2\} \big\},
\end{equation}

For each point $x'_j$ in the second image, $D_2(j) \subset \{1...K_2\}$  is defined similarly and the accumulator entries $(j,m,n)$ are incremented according to 

\begin{equation}
\big\{(j,m,n)| m \in \{1...K_1\} , n \in D_2(j) \big\},
\end{equation}

The pair of matching lines with the maximum number of putative correspondences are given by  $(l_{{\bf m}^*},l_{{\bf n}^*}')$ where 

\begin{equation}
({\bf m}^*,{\bf n}^*) =\argmax_{m,n} \sum_{j} A(j,m,n),
\end{equation}

In case there is more than a single line we randomly select one. 

\section{Separable Four Points Fundamental Matrix}
\label{Sec:Sep}

Given a set of putative correspondences $\{(x_i,x_i')\}_{i=1..N}$, we compute the fundamental matrix by two steps. Our computation is based on RANSAC. We assume inlier ratio $R$, epipolar homography inlier threshold $T_1$ and fundamental matrix inlier threshold $T_2$.\\

{\bf Step One.}  Based on Sec.~\ref{Sec:Lines} , we recover $(l_{{\bf m}^*},l_{{\bf n}^*})$, which are the corresponding image lines with the maximum number of $K$ putative corresponding points ${(x_i,x_i')}_{1..K}$.  We select fixed control points for each line  $l_{{\bf m}^*}$ and $l_{{\bf n}^*}$. The points are selected on the intersection of the lines and the boundaries of the images. For each pair of putative corresponding points on the lines, we compute their representation by the control points with the two possible orientations and use the second one only if the first fails. We use RANSAC to recover the epipolar homography. We iteratively sample three putative corresponding points, compute the epipolar homography, transfer each point on the line to the corresponding line, and count the number of inliers based on the Euclidean distance and the threshold $T_1$. The number of required iterations for the three parameter model is often small. For example, for an outlier rate of 60\%, we need only 71 iterations to recover the correct model with a confidence of 99\%. Note that although we sample three points, we require at least four corresponding points on line segment; the additional points are used for validation of the hypothesized epipolar homography.\\

{\bf Step Two.} Given the three points selected, we iteratively sample four additional points and compute the fundamental matrix based on the seven points algorithm. The number of iterations is computed based on $R$ with respect to four parameters. We count the number of inliers based on $T_2$ with respect to the symmetric epipolar distance \cite{hartley2003multiple}. For the same case as above of an outlier rate of 60\%, for the four parameters' model we need only 178 iterations.\\

The above refers to the implementation of our approach based on the seven points algorithm. For implementation based on the eight points algorithm, we sample five points instead of four in the second step.


\section{Experiments}
\label{Sec:exp}

\subsection{Datasets}
We conduct experiments on the following real world datasets: (1) The Strecha datasets \cite{strecha2008benchmarking} which include ground truth cameras and reconstruction, (2) The Tower of London sequence from the Flickr dataset \cite{wilson2014robust}, where the images were downloaded from Flickr based on geotag. It includes images captured by the community at random camera positions with a large variety of urban outdoor scenes, and (3) The Family sequence from the Tanks and Temples dataset \cite{Knapitsch2017}, which includes wide baseline camera poses, with medium sized images captured by a hand-held camera in an urban yard scene.

For the last two datasets, we sampled 2000 image pairs with at least 20 matches with a symmetric epipolar distance of less than or equal to 1 pixel with respect to the ground truth, and used them for the evaluation. We show that our approach is able to successfully compute the fundamental matrices for these real-world diverse sets of image pairs, leading to an efficient estimation process. Table~\ref{table:datasets} shows the properties of the datasets.

\begin{table}[t]
\begin{center}
\begin{tabular}{l|c|c}

 & Pairs & Ground truth \\
\hline\hline
Strecha & 569  & Given \\
Flickr & 2000 & COLMAP\\
Tanks and Temples & 2000 & COLMAP\\

\end{tabular}
\end{center}
\caption{The datasets. For both Flickr and Tanks and Temples we obtained the ground truth fundamental matrices based on COLMAP reconstruction.}
\label{table:datasets}
\end{table}

\begin{figure}[t]
\begin{center}
   \includegraphics[width=0.88\linewidth]{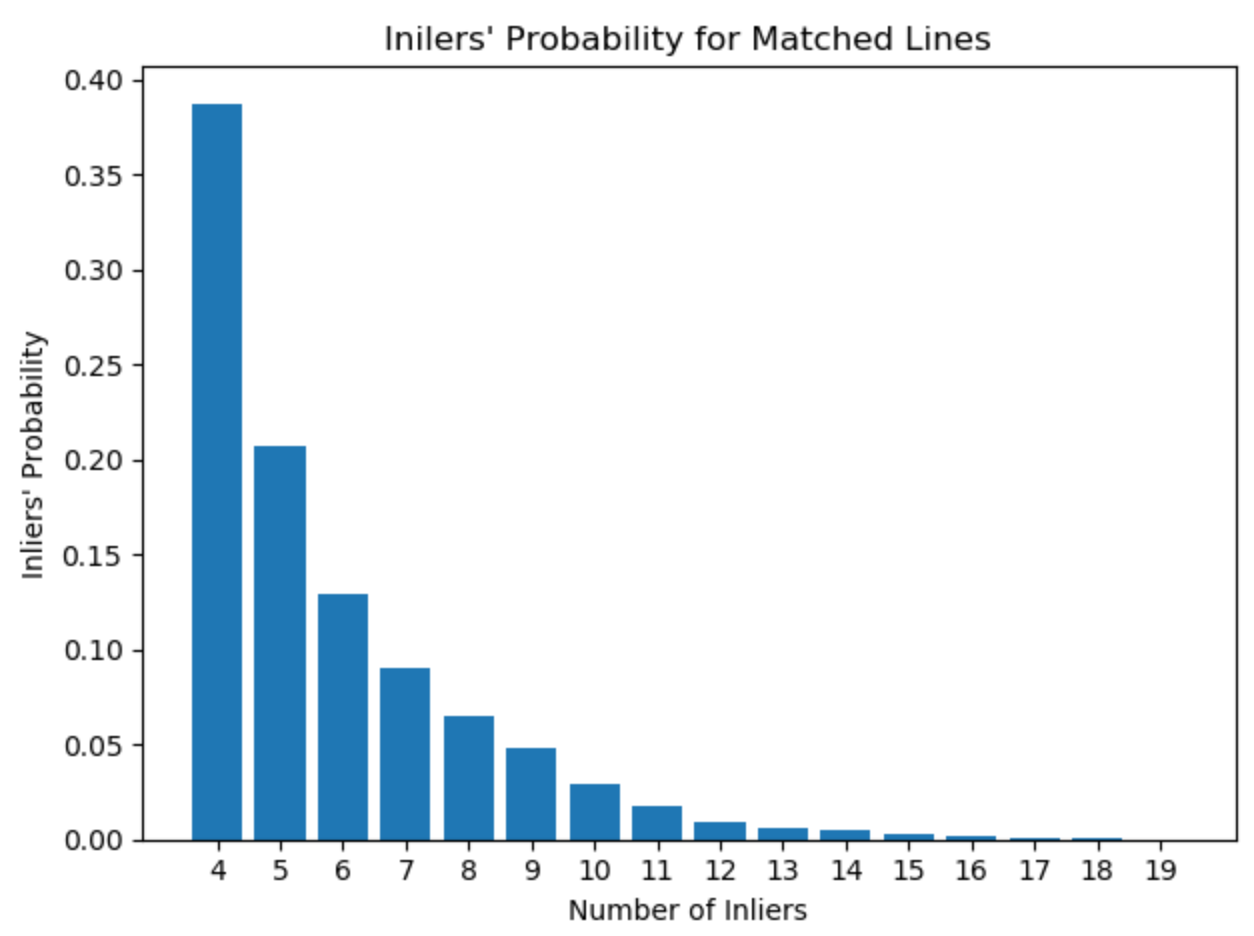}
\end{center}
   \caption{The probability for a given number of inliers in matched lines. The x-axis is the number of inliers and the y-axis is the probability. The average number of inliers is 5.83.}
\label{fig:line_inliers}
\end{figure}
\begin{figure}[t]
\begin{center}
   \includegraphics[width=0.88\linewidth]{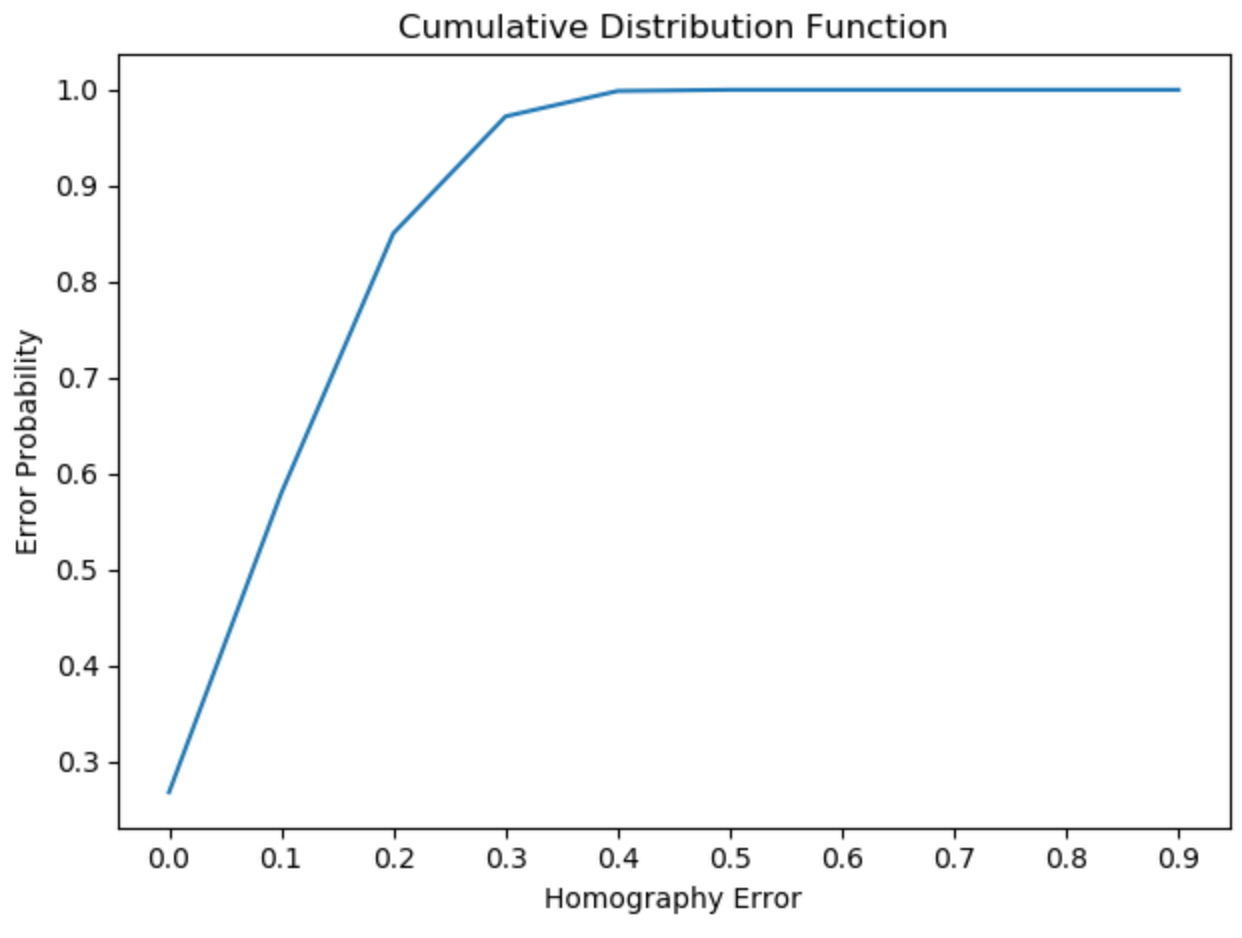}
\end{center}
   \caption{The cumulative distribution function (CDF) of the line homography error in matched lines. The x-axis is the error in pixels and the y-axis is the probability. 99\% of the inliers are with a homography error of equal to or less than $0.5$.}
\label{fig:CDF_homography}
\end{figure}

\subsection{Matching Co-Linear Points}
In this subsection, we discuss the matching process of corresponding lines with at least four points, based on the Strecha dataset.

{\bf How Many Lines?} On average, the Hough transform recovers 286.4 independent lines in each image. Out of all possible pairs of matched lines, the average number of lines with corresponding points across a pair of images is 3354.91, including lines with outliers. 45.8\% (1538.84) of the matched lines are lines with at least four correspondences, where a corresponding point might be either inlier or outlier.

{\bf Quality of Lines}. Overall, we would like to match lines with a high number of inliers and a low homography error. The average number of points, including outliers, on matched lines is 7.62. The average number of inliers on corresponding lines is 5.83, for a homography threshold of 1. Fig.~\ref{fig:line_inliers} shows the probability of having a given number of inliers in matched lines, for such a homography error. Fig.~\ref{fig:CDF_homography} presents the probability for a given homography error on the matched lines. The homography error of 97\% of the inliers is below $0.4$ and 99\% are below 0.5, which is used as our default threshold.

\begin{figure}[t]
\begin{center}
   \includegraphics[width=0.475\linewidth]{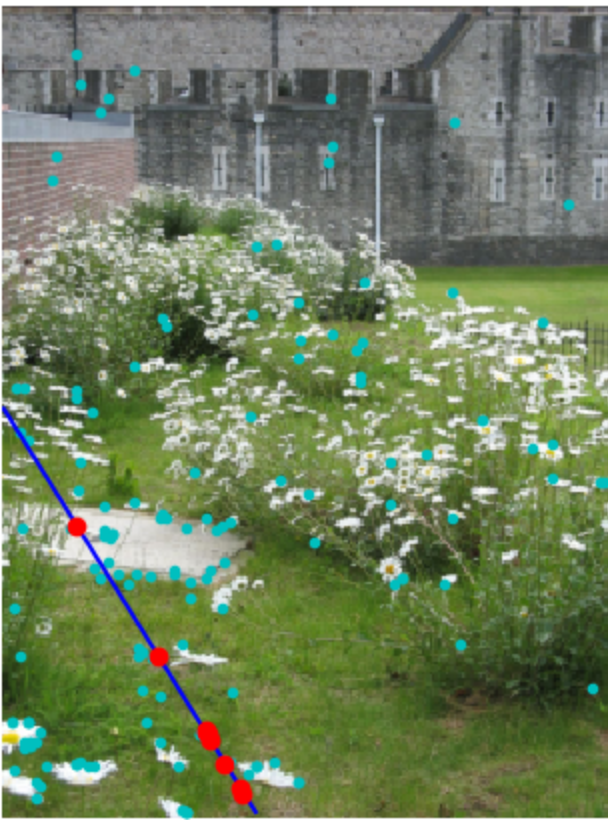}
   \includegraphics[width=0.48\linewidth]{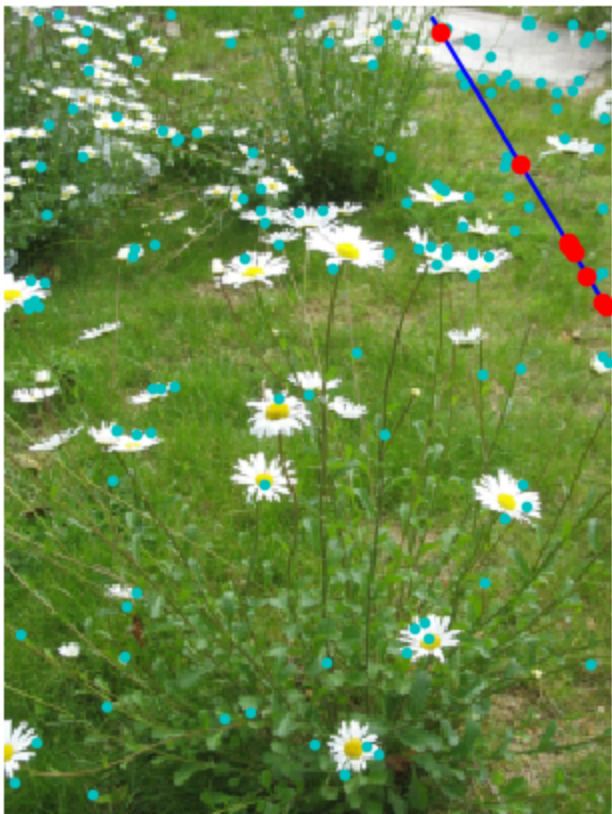}\\
    \includegraphics[width=0.48\linewidth]{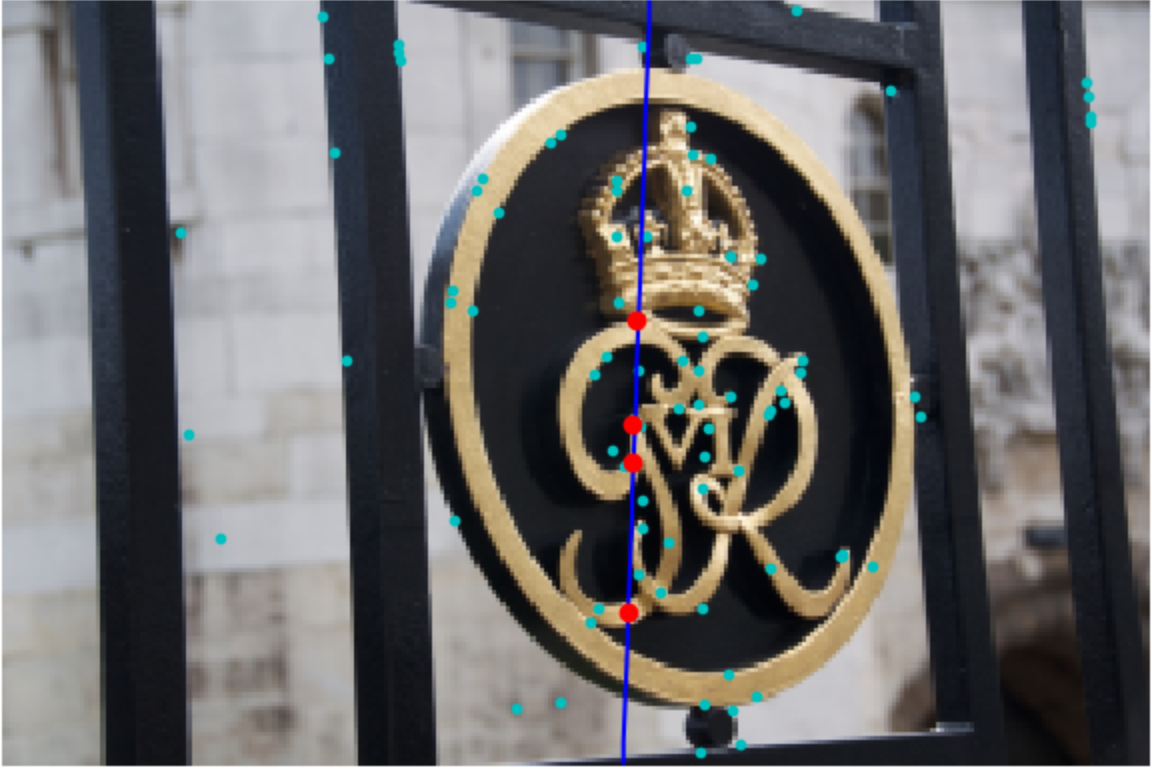}
   \includegraphics[width=0.48\linewidth]{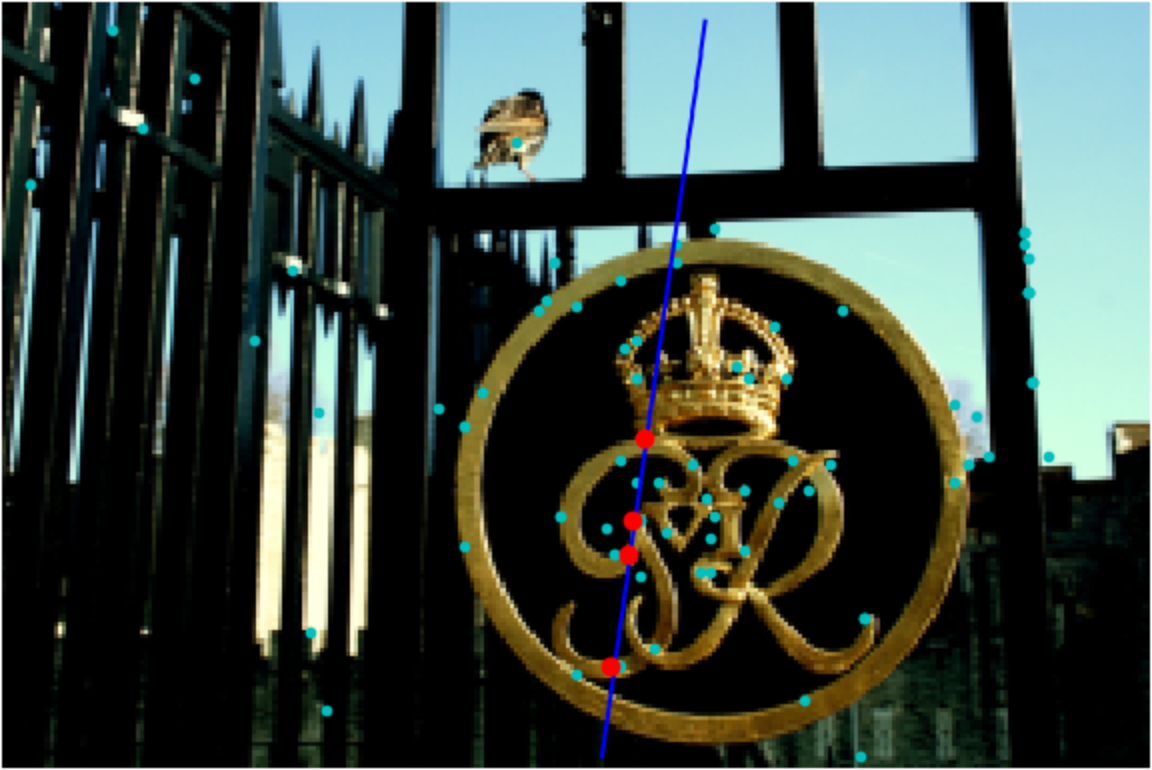}\\
   \includegraphics[width=0.48\linewidth]{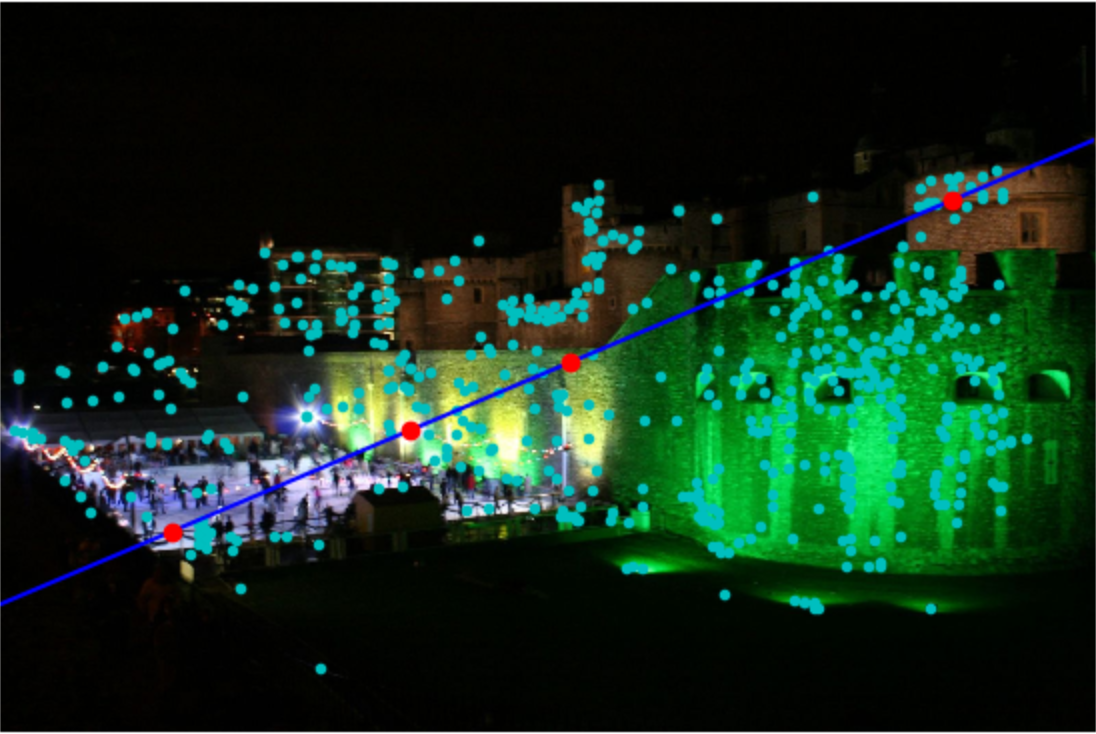}
   \includegraphics[width=0.48\linewidth]{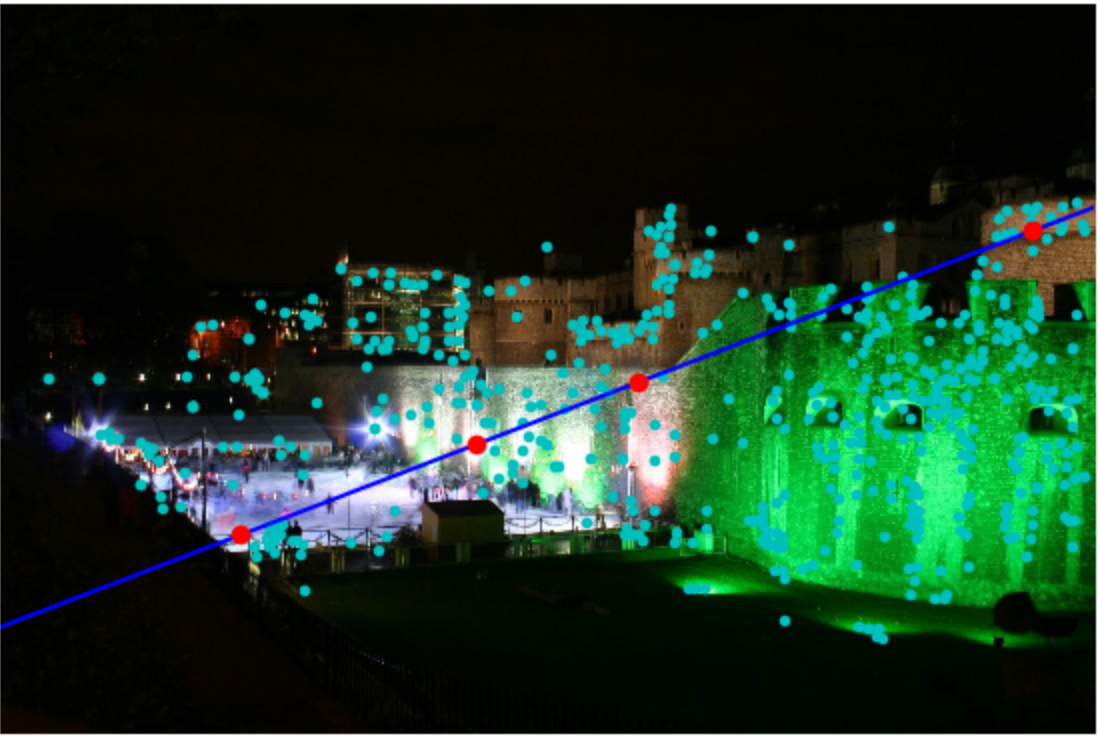}\\
\end{center}
   \caption{Examples of matched lines recovered by our approach for various scenes from Tower of London dataset. The red circles are the matched points on the corresponding lines and the cyan dots are the putative correspondences. The inliers ratio are  $0.18,0.28$ and $0.85$ from top to bottom. }
\label{fig:line_matches}
\end{figure}

{\bf Scene Structure}. Our approach often successfully matches lines in both near and far scenes and low and high inlier rates.  Fig.~\ref{fig:line_matches} shows examples from Tower of London dataset of matched lines and corresponding points recovered by our approach, with inlier ratios of $0.18,0.28$ and $0.85$, for various scenes. For the first and second row from the top, there are 27 and 25  inliers out of 148 and 89 putative correspondences, respectively. For the last row, there are 498 inliers out of 582 putative correspondences.

\begin{figure}[t]
\begin{center}
   \includegraphics[width=0.48\linewidth]{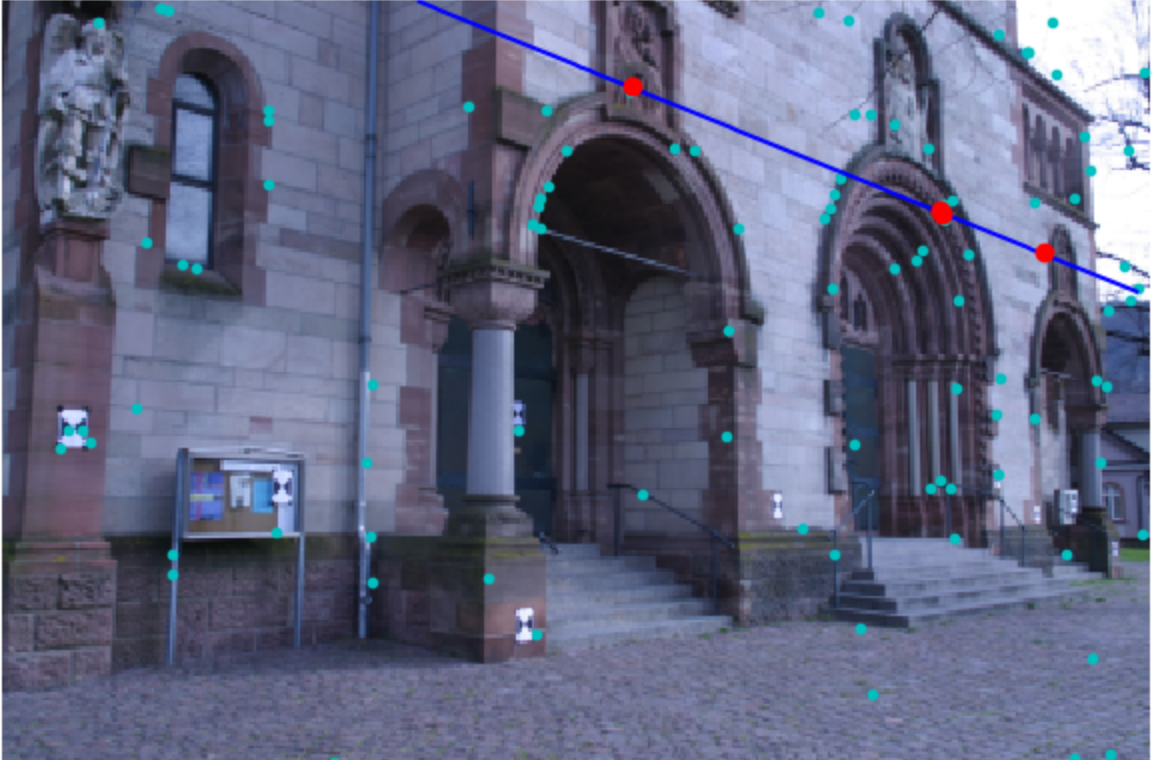}
   \includegraphics[width=0.48\linewidth]{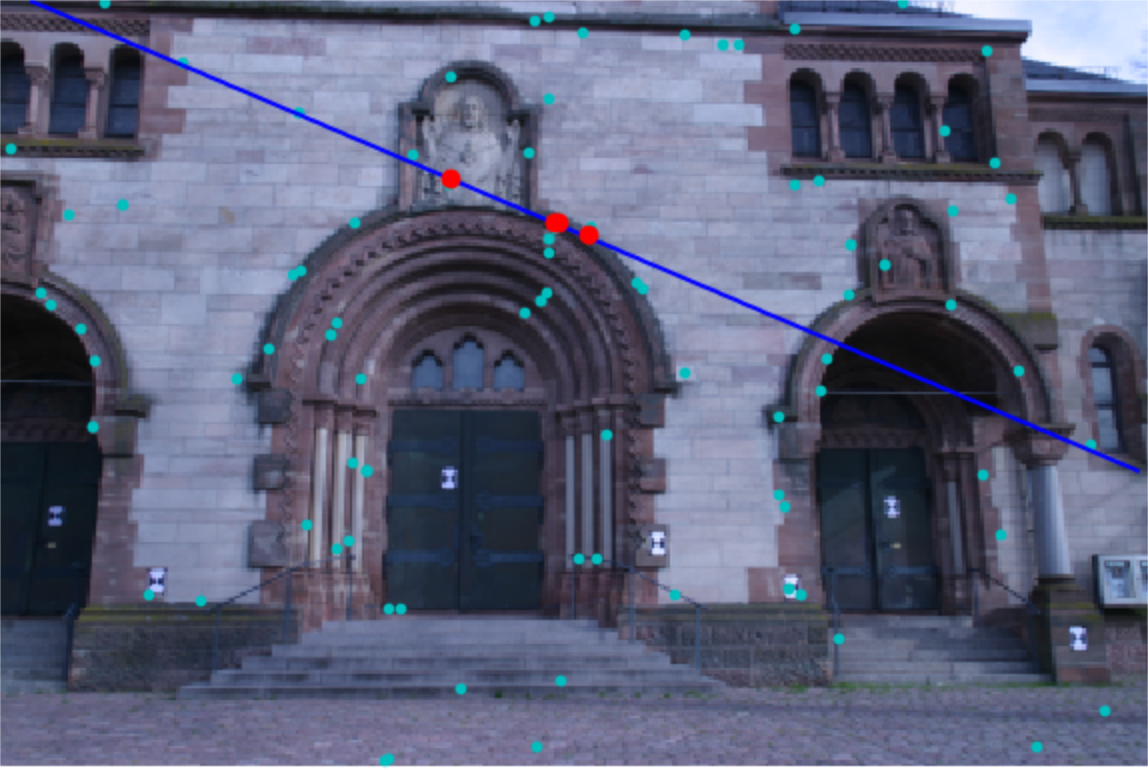}\\
\end{center}
   \caption{An example of a failure case. The the best possible corresponding lines are with wrong matches and the homography could not be recovered. }
\label{fig:line_failure}
\end{figure}

{\bf Missing Inliers}  Relative to RANSAC, our approach failed to recover at least 20 inliers across images in 4.9\% of the cases. Qualitatively, these cases occur when there is only a small number of inliers, both in the images and on the recovered lines. For the failure cases, the average number of inliers is 26.8 and the maximum is 34. Quantitative analysis indicates that in such cases the best possible matching of corresponding lines across images contains wrong or inaccurate matches and the homography could not be recovered. Fig.~\ref{fig:line_failure} shows such an example. The overall percentages of failure cases over all datasets are presented in Table~\ref{table:failure_all}.

\subsection{Evaluation}
\label{subsec:eval}

{\bf Methods}. We compared RANSAC-based~\cite{Fischler} and LMEDS-based~\cite{rousseeuw1984least} computations of the fundamental matrix based on the following minimal solvers' methods:
\begin{itemize}
    \item RANSAC with the standard eight points algorithm, denoted as RANSAC.
    \item LMEDS with the standard eight points algorithm, denoted as LMEDS.
    \item State-of-the-art minimal solver of \cite{barath2018five} combined with GC-RANSAC \cite{barath2018graph}, which is based on homography and rotation information of five points, denoted as Hom-Rot-5.
    \item RANSAC with our minimal sampling solver, using the seven points algorithm based on two samples of three and four points, denoted as Ours-RANSAC-4.
    \item RANSAC with our minimal sampling solver, using the eight points algorithm based on two samples of three and five points, denoted as Ours-RANSAC-5.
    \item LMEDS with our minimal sampling solver, using the seven points algorithm based on two samples of three and four points, denoted as Ours-LMEDS-4.
    \item LMEDS with our minimal sampling solver, using the eight points algorithm based on two samples of three and five points, denoted as Ours-LMEDS-5.
\end{itemize}

{\bf Methodology}. For the baseline methods we performed a grid-search over hyperparameters. For sequences with no ground truth fundamental matrices, we reconstructed the sequences using COLMAP \cite{schoenberger2016sfm}. COLMAP is a general purpose Structure-from-Motion (SfM) pipeline which globally infers the 3D structure, leading to accurate results. We obtained the camera poses and fundamental matrices, and used them as the ground truth. 

For all sequences, we used the SIFT \cite{lowe2004distinctive} descriptor to extract and match putative correspondences between all pairs of images. We used the ground truth fundamental matrices to evaluate the symmetric epipolar distance of the putative matching. Each image pair with less than 20 matches within a symmetric epipolar distance of one is discarded.

{\bf Metrics}.  We report the average symmetric epipolar distance resulting from our approach and the baselines. Unless otherwise stated, we used a threshold of 3 pixels in our approach as the inlier threshold and $0.5$ as the epipolar homography threshold. We use a fixed homography’s threshold over all the datasets, demonstrating its robustness. We report the percentage of inliers found by each method, and report the F-measure, where corresponding points with symmetric epipolar distance less than 1 pixel with respect to the ground truth fundamental matrix is considered as positive.

\subsection{Accuracy}

\begin{table}[tb]
 \footnotesize
\begin{center}
\begin{tabular}{ l|c|c|c }  
\hline
  &             \%~inliers      & F-Score         &   Mean       \\ \hline
RANSAC           &  71.2        & 0.89 $\pm$ 0.05            &   0.61       \\ 
LMEDS            &  71.1        & 0.88  $\pm$ 0.08          &    {\bf 0.57}      \\ 
Hom-Rot-5        &  71.5        & 0.90 $\pm$ 0.06           &   0.60      \\ \hline
Ours-LMEDS-4     &  62.4        & 0.81  $\pm$ 0.06           &   0.72      \\ 
Ours-LMEDS-5     & 72.1         & 0.90   $\pm$ 0.05          &   0.59      \\ 
Ours-RANSAC-4    &  61.7        & 0.82   $\pm$ 0.06          &   0.71      \\ 
Ours-RANSAC-5    & {\bf 72.2} & {\bf 0.91 $\pm$ 0.05 }     &  0.58     \\ 
\hline
\end{tabular}
\end{center}
\caption{Results of the Strecha dataset, based on the ground truth camera poses.
\label{table:accuracy_strecha}}
\end{table}

\begin{table}[tb]
 \footnotesize
\begin{center}
\begin{tabular}{ l|c|c|c }  
\hline
              &\%~inliers & F-Score & Mean    \\ \hline
RANSAC         &  43.7 & 0.52 $\pm$ 0.08 &   0.63      \\ 
LMEDS           &  43.2 & 0.53  $\pm$ 0.09 &   0.69     \\ 
Hom-Rot-5       &  44.5 & 0.58  $\pm$ 0.08 &   0.61     \\ \hline
Ours-LMEDS-4  &  41.6 & 0.51 $\pm$ 0.1 &   0.74      \\ 
Ours-LMEDS-5 &  46.5 & 0.60  $\pm$ 0.09 &   0.57      \\ 
Ours-RANSAC-4  &  42.9 & 0.51 $\pm$ 0.1 &   0.78      \\  
Ours-RANSAC-5  & {\bf  49.3} & {\bf 0.62 $\pm$ 0.09} &  {\bf 0.54}     \\ 
\hline
\end{tabular}
\end{center}
\caption{Results of Tower of London sequence from the Flickr dataset. 
\label{table:accuracy_flickr}}
\end{table}

\begin{table}[tb]
 \footnotesize
\begin{center}
\begin{tabular}{ l|c|c|c }  
\hline
                 & \%~inliers   & F-Score & Mean    \\ \hline
RANSAC           &  45.4 & 0.72  $\pm$ 0.07 &   0.69             \\ 
LMEDS            &  44.2 & 0.71 $\pm$ 0.08 &   0.68              \\ 
Hom-Rot-5        &  45.5 & {\bf0.78  $\pm$ 0.07} &  {\bf 0.61}          \\ \hline
Ours-LMEDS-4   &  43.8 & 0.61  $\pm$ 0.09 &   0.71              \\ 
Ours-LMEDS-5   &  45.7 & 0.71  $\pm$ 0.07 &   0.66              \\ 
Ours-RANSAC-4  &  44   & 0.62  $\pm$ 0.08 &   0.69           \\ 
Ours-RANSAC-5  &  {\bf 48.1} &  0.73  $\pm$ 0.07 & 0.63       \\ 
\hline
\end{tabular}
\end{center}
\caption{Results of the Family sequence from the Tanks and Temples dataset. 
\label{table:accuracy_tanks}}
\end{table}

{\bf Strecha.} The dataset contains eight multi-view collections of high-resolution images ($3072 \times 2048$), provided with ground truth camera poses. For the computation of matching lines with putative corresponding points across images, we sampled $150$ putative correspondences, and used these lines with respect to all existing points' correspondences. We used SIFT ratio-test of $0.75$. The binary images used for matching the lines  (Sec.~\ref{Sec:Lines}) are of a lower resolution, as suggested in~\cite{tola2009daisy,trulls2013dense}, and the points coordinates are resized accordingly. Unless noted otherwise, the width resolution of $512$ was used for the line matching phase with the original aspect ratio. The reported results are with respect to the original resolutions. The sequences are of images with varying distances between focal points with a total of 569 valid image pairs.  As suggested in~\cite{barath2018five}, we used the five-points' solver of \cite{barath2018five} within the GC-RANSAC \cite{barath2018graph} estimator. Table.~\ref{table:accuracy_strecha} presents our results. In both \%-inliers and F-Score metrics our RANSAC-based eight points approach outperforms the baselines.

{\bf Flickr.} The dataset obtained by downloading images of specific scenes from Flickr represents the most challenging case. The images captured random scenes at different times, with various cameras from very different positions. We use the Tower of London sequence for comparison. We reconstruct the sequence using COLMAP\cite{schoenberger2016sfm} and use the reconstruction as ground truth. The results are presented in Table~\ref{table:accuracy_flickr}, where it can be seen that our RANSAC-based eight points approach clearly outperforms all other methods.

{\bf Tanks and Temples.} The dataset includes medium-resolution images ($1920 \times 1080$). For comparison, we used the Family sequence. As ground truth, we use the COLMAP Reconstruction. As shown in Table~\ref{table:accuracy_tanks}, our RANSAC-based eight points approach recovered the highest number of inliers, with mean error similar to~\cite{barath2018five}. Both outperform the baselines. \\

\begin{figure}[t]
\begin{center}
\includegraphics[height=0.75\linewidth]{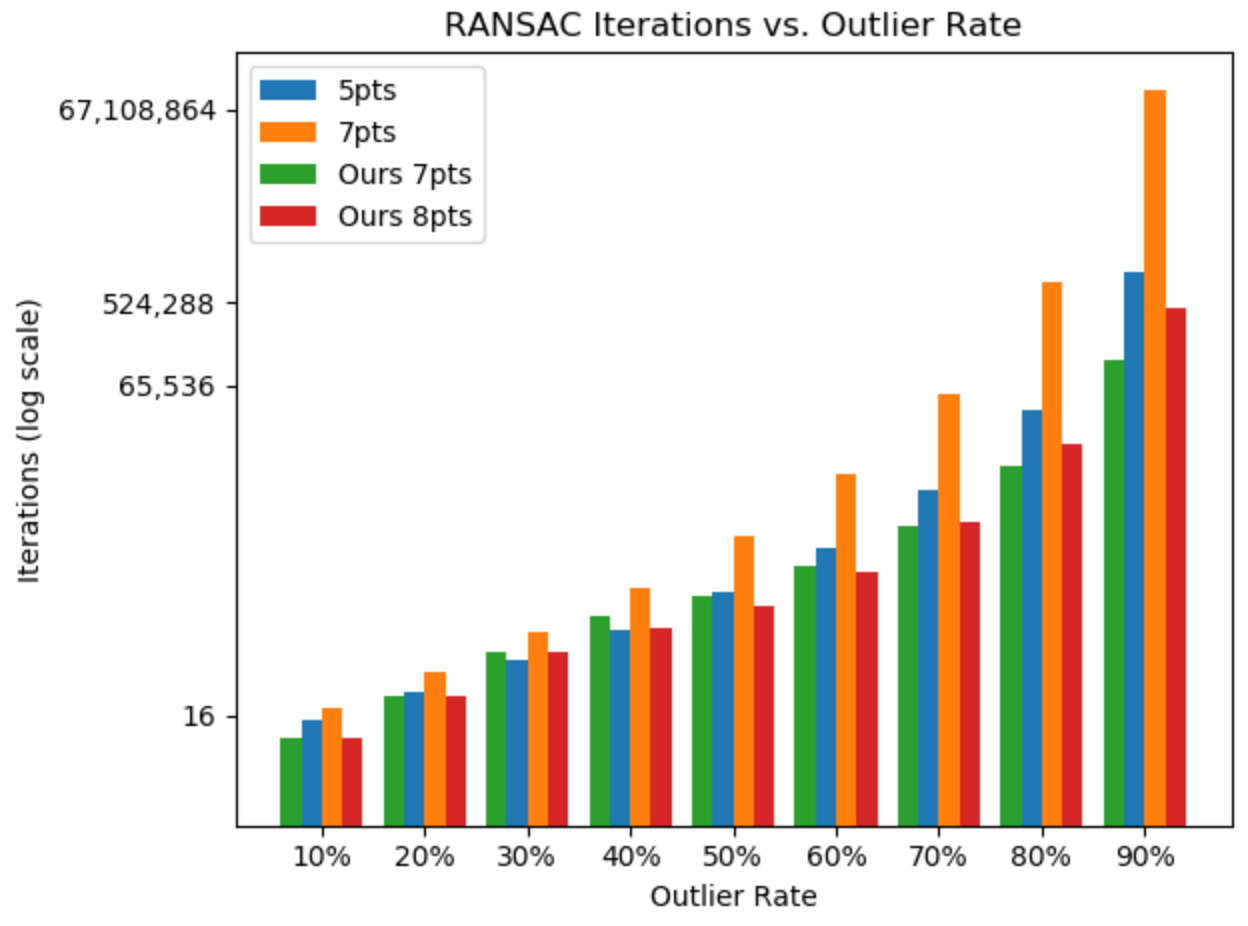}\\
\end{center}
   \caption{The expected number of RANSAC iterations based on the actual number of solutions per sample in the 7-point algorithm and the additional preprocessing iterations required by our approach.} 
\label{fig:ransac_iterations_practical}
\end{figure}

\begin{table}[tb]
 \footnotesize
\begin{center}
\begin{tabular}{ l|c|c }  
\hline
  & RANSAC Iterations & \%Failure     \\ \hline
RANSAC       &  2931 &  -  \\ 
Hom-Rot-5       &  671 &  4.3  \\ 
Ours-RANSAC-4  &  {\bf 627}  & 7.2  \\ 
Ours-RANSAC-5  &  643  &  {5.1}  \\ 
\hline
\end{tabular}
\end{center}
\caption{The number of RANSAC iterations and failure cases relative to RANSAC for all three datasets. 
\label{table:failure_all}}
\end{table}

\subsection{Efficiency}
\label{sec:eff}

We measured the number of RANSAC iterations required by each method and the number of failure cases.  For the total number of RANSAC iterations, we also consider all validation iterations. For methods which are based on the seven points algorithm, both ours and the baseline, there might be more iterations than the theoretical number of required samples.  A failure is considered if the solver has not been able to accurately recover at least 20 matches, in case there are indeed such ground truth matches.  

The solver of \cite{barath2018five} is combined within another robust estimator framework, the GC-RANSAC \cite{barath2018graph}. We execute only the solver itself with the optimal number of iterations as an initialization.  Our approach starts by finding the matching lines with the maximum number of putative correspondences (Sec.~\ref{Sec:Sep}).  Based on our experiments the runtime of our lines' matching step is equivalent to 71 RANSAC iterations and the number of solutions per sample for the 7-point algorithm is 2.43. To reduce the additional overhead associated with our matching step we use instead the standard eight points' RANSAC for inlier ratio less than $0.3$. For such inlier ratio the 8-point algorithm requires only one hypothesis' evaluation per sample, with overall $9$-$78$ iterations instead of $20$-$132$ for the 7-point algorithm. Fig.~\ref{fig:ransac_iterations_practical} presents the required number of iterations by the different approaches based on the actual number of solutions per sample including the additional iterations required by our matching lines phase.

Table~\ref{table:failure_all} shows the percentage of cases that failed for each approach and the average number of iterations, over all the three datasets, the Strecha, Flickr and Tanks and Temples.

\section{Conclusion}
We presented a novel approach for the computation of the fundamental matrix based on epipolar homography decomposition. Our approach can reduce, both theoretically and practically, the number of required iterations to a new minimal number.  The standard RANSAC procedure can be incorporated into our approach, providing an overall robust and efficient solution which is well suited for Structure from Motion (SfM) pipelines.

{\small
\bibliographystyle{ieee_fullname}
\bibliography{egbib}
}

\end{document}